# Scalable Feature Subset Selection for Big Data using Parallel Hybrid Evolutionary Algorithm based Wrapper in Apache Spark

Yelleti Vivek[1,2], Vadlamani Ravi[1*] and P. Radhakrishna[2]

[1]Center of Excellence in Analytics,
Institute for Development and Research in Banking Technology,
Castle Hills Road #1, Masab Tank, Hyderabad-500076, India
[2]Department of Computer Science and Engineering, National Institute of Technology,
Warangal-506004, India
yvivek@idrbt.ac.in; padmarav@idrbt.ac.in; prkrishna@nitw.ac.in

**Abstract**

Owing to the emergence of large datasets, applying current sequential wrapper-based feature subset selection (FSS) algorithms increases the complexity. This limitation motivated us to propose a wrapper for feature subset selection (FSS) based on parallel and distributed hybrid evolutionary algorithms (EAs) under the Apache Spark environment. The hybrid EAs are based on the BDE and Binary Threshold Accepting (BTA), a point-based EA, which is invoked to enhance the search capability and avoid premature convergence of the PB-DE. Thus, we designed the hybrid variants (i) parallel binary differential evolution and threshold accepting (PB-DETA), where DE and TA work in tandem in every iteration, and (ii) parallel binary threshold accepting and differential evolution (PB-TADE), where TA and DE work in tandem in every iteration under the Apache Spark environment. Both PB-DETA and PB-TADE are compared with the baseline, viz., the parallel version of the binary differential evolution (PB-DE).  All three proposed approaches use logistic regression (LR) to compute the fitness function, namely, the area under ROC curve (AUC). The effectiveness of the proposed algorithms is tested over the five large datasets of varying feature space dimension, taken from cyber security and biology domains. It is noteworthy that the PB-TADE turned out to be statistically significant compared to PB-DE and PB-DETA. We reported the speedup analysis, average AUC obtained by the most repeated feature subset, feature subset with high AUC and least cardinality.

## 1. Introduction

Selecting relevant and important feature subset is evidently a paramount pre-processing step in the CRISP-DM methodology [72] of data mining. This process of selecting the important group of features is popularly known as Feature subset selection (FSS) [1,2]. The main objective of FSS is to select the most relevant and highly discriminative feature subsets. The spectacular benefits of FSS are as follows: it improves the comprehensibility of the models, reduces the model complexity, improves the training time, avoids overfitting, and sometimes improves the model's performance. Further, the resulting model becomes parsimonious. The ubiquitous presence of big datasets in every domain made FSS a mandatory pre-processing step.

FSS can be performed primarily in three different ways: filter, wrapper, embedded approaches. The main difference lies in the fitness value determination and selecting the salient features either individually or group-wise. The filter approaches measure the fitness value based on the statistical measure such as Information gain, Mutual information, Gain ratio, etc. These methods are fast but result in less accuracy and cannot account for the interaction effects amongst the features. Wrapper approaches comprise a metaheuristic optimization algorithm that searches for the best feature subsets as indicated by the highest fitness value determined by a classifier (for a classification problem) or a regression model (for a regression problem). These approaches are computationally intensive but highly accurate while accounting for the interaction effects of the features. In Embedded approaches, the feature subset selection is embedded as a part of the model-building phase. These approaches combine the advantages of being less computationally expensive than wrapper and give better accuracy than filter approaches. Even though wrapper approaches impose high complexity, the selected feature subset is highly generalized to the underlying classifier.

---

* Corresponding Author, Phone: +914023294310; FAX: +914023535157

In current study, FSS is formulated as a combinatorial problem because if there are *n* features, the total possible number of feature subsets is $2^n - 1$. Accordingly, the total number of feature subsets that can be formed constitutes the search space. Now, the objective is to search for an efficient feature subset which comprises less redundant features. The best feature subset is found out by checking all the possible feature subsets. However, this is a brute force method and becomes unwieldy when the feature space dimensions *n* becomes large, as in big datasets. Metaheusristics (evolutionary algorithms (EAs) subsumed) have demonstrated their superiority over conventional optimization methods in solving various combinatorial and continuous optimization problems. Metaheuristics are of two different types: (i) point-search-based methods such as Threshold Accepting (TA), Simulated Annealing (SA) etc., (ii) EAs, which are population-based EAs such as genetic algorithgm (GA), differential evolution (DE), etc. The present research study posed the FSS in a single objective environment where the objective function maximizes the AUC, thereby selecting the feature subsets of length less than or equal to *k* (where *k* < *n*) while achieving the best possible AUC.

As the data is generated in large volumes at a phenomenal rate, scalability becomes a major concern while developing solutions to analyse such big data. Therefore, designing scalable solutions gained its prominence. MapReduce [70] is a programming paradigm used in handling such big data. It mainly consists of two steps: map and reduce. MapReduce solutions are proven to be scalable. There are different big data frameworks available to design MapReduce solutions. Among them Apache Spark is faster in nature due to its in-memory computation feature. Apache Spark is an open-source, fast computing distributed engine used to handle such large amounts of data. Spark uses in-memory computing by using Resilient Distributed Datasets (RDD) that boost the performance, thereby avoiding the disk-access. RDD is inherently distributed in nature, follows the lazy evaluation and is immutable in nature. Apache Spark also provides versatility by leveraging to combine with other big data tools such as Hadoop.

Extant EA -based wrapper algorithms are sequential and limited to small datasets. Even though they can be applied on larger datasets, they perform poorly. In the current world, the generation of datasets is growing by leaps and folds, thereby demanding the development of scalable wrappers for FSS. There is a growing need to develop such parallel wrappers for FSS in the context of big data. This motivated us to propose a scalable wrapper for FSS in single objective environment. In the current study, the objective function is to select the feature subset of length less than or equal to *k* (where *k* < *n*) while achieving the best possible AUC. To the best of our knowledge, no work is reported so far, where feature subset selection is either performed by a scalable wrapper involving parallel and distributed EC techniques or their hybrids under the Apache Spark environment.

This paper's significant contributions are as follows: (i) Parallel DE is designed under Apache Spark to develop wrappers for FSS. (ii) Binary versions of the TADE and DETA are developed and parallelized under Apache Spark. These are named as PB-TADE and PB-DETA respectively. (iii) Then, these are invoked to develop wrappers for FSS, where logistic regression is chosen as the classifier to evaluate the fitness function, namely the AUC. (iv) To achieve scalability and algorithm parallelization, we proposed a novel MapReduce-multithread based framework.

The remainder of the paper is structured as follows: Section 2 presents the literature review. Section 3 presents an overview of TA and DE. Section 4 presents the proposed methodology. Section 5 describes the datasets and experimental setup. Section 6 discusses the results obtained by the models. Finally, section 7 concludes the paper.

## 2. Literature Review

Differential evolution, one of the widely used algorithms for the feature subset selection, is proposed by Storn and Price [3]. Table 1 presents the details the filter, wrapper sequential versions of DE, where feature selection is posed as a combinatorial optimization problem. Zhang et al.[4] proposed a modified DE with self-learning (MOFS-BDE). In [4], authors had introduced three different operators, namely: (i) modified binary mutation operator based on the probability difference, (ii) one bit purifying search operator (OPS) to improve the self-learning capability of the elite individuals, and (iii) non-dominated sorting in the selection phase to reduce the computational complexity involved in the selection operator. Vivekanandan and Iyengar [5] designed a two phase solution, where the critical features are selected by the DE and fed into the integrated model of the feed-forward neural network and fuzzy analytic hierarchy process (AHP) [6]. Nayak et al.[7], proposed FAEMODE, a filter approach using elitism based multi-objective DE. It can handle both the linear and non-linear dependency among the features via both the correlation coefficient (PCC) and mutual information (MI). Mlakar et al. [8] designed a multi-objective DE (DEMO) based wrapper for facial recognition systems. In their approach, initially, the important features are extracted based on the histogram of oriented gradient descriptor (HOG) and fed to the DEMO to find the pareto optimal solutions. Khushaba et al. [9] proposed DE with a statistical repair mechanism, DEFS, for selecting the optimal feature subsets in datasets with varying dimensionality. In their proposal, the probability of the feature distribution is fed to DEFS by the roulette wheel. Hancer et al. [10] proposed filter-based DE, MIFS, where the fisher score determines the mutual relevance between the features and class labels. The features are assigned a rank based on their fisher score. Hancer [11] proposed a new multi-objective differential evolution (MODE-CFS) with two stage mutation, (i) centroid-based mutation to perform clustering and (ii) feature-based mutation to perform feature selection. Non-dominated sorting is applied to determine the pareto optimal solution set. Ghosh et al. [12] proposed self-adaptive DE (SADE) based wrapper for the feature selection in the hyperspectral image data. The selected feature subsets are fed into fuzzy-KNN to obtain accuracy. Bhadra and Bandyopadhyay [13] improved the modified DE. They proposed an unsupervised feature selection approach called MoDE with the objective functions as (i) average dissimilarity of the selected feature subset, (ii) average similarity of the non-selected feature subset, and (iii) the average standard deviation of the selected feature subset. All the objectives mentioned above use normalized mutual information. Baig et.al.[14] proposed modified DE based wrapper for the Motor Imagery EEG having a high dimensional dataset. SVM is used here as a classifier. Almasoudy et al. [15] designed a wrapper feature selection based on modified DE. The authors considered Extreme Learning Machine (ELM) as a classifier and tested its effectiveness over the intrusion detection dataset NSL-KDD. Zorarpaci and Ozel [16] proposed a hybrid FS approach based on DE and Ant-Bee Colony (ABC), where the J48 classifier from Weka computes the fitness score. This hybrid model achieved a significant F-score with less cardinal feature subset than the stand-alone DE and stand-alone ABC. Then, a quantum-inspired wrapper based on DE (QDE) with logistic regression as the classifier was proposed by Srikrishna et al. [17]. They reported that QDE achieved better repeatability than the BDE. Lopez et al. [18] proposed a wrapper based on permutation DE, where the permutation-based mutation replaced the mutation operator, and the diversity of the generated children solutions is controlled by using a modified recombination operator. Zhao et al. [19] developed a two-stage wrapper feature selection algorithm, where in the first stage, the fisher score and information gain are applied to filter the redundant features. Then in the second stage, the top-k features are passed to the modified DE to perform the feature selection on four different breast cancer datasets. Hancer [20], for the first time, used fuzzy and kernel measures as filters to calculate the mutual relevance and redundancy with DE to handle continuous datasets. Li et al.[21] designed DE-SVM-FS and compared it with the default SVM approach, and they demonstrated that the DE and SVM-based FS achieved better accuracy than the stand-alone SVM. Wang et al.[22] proposed DE-KNN, where the KNN is the classifier. DE-KNN performed both the feature selection as well as the instance selection.

Table 1: Sequential versions of DE and its wrapper variants

| Authors | # Objectives | Algorithm | Wrapper (classifier) / Filter |
|---|---|---|---|
| Zhang et.al. [4] | Multi Objective | Self-Learning DE | Wrapper (KNN) |
| Vivekanandan and Sriraman [5] | Single Objective | Modified DE | Filter |
| Nayak et.al. [7] | Multi Objective | FAEMODE | Filter |
| Mlakar et.al. [8] | Multi Objective | DE+HOG | Wrapper (SVM) |
| Khushaba et.al.[9] | Single Objective | DEFS | Filter |
| Hancer [11] | Multi Objective | MODE-CFS | Filter |
| Hancer et.al. [10] | Multi Objective | DE + MIFS | Filter |
| Ghosh et.al. [12] | Multi Objective | SADE | Wrapper (Fuzzy-KNN) |
| Bhadra and Bandyopadhyay [13] | Multi Objective | MoDE | Filter MI |
| Bhaig [14] | Multi Objective | Modified DE | Wrapper (SVM) |
| Almasoudy et.al. [15] | Multi Objective | Modified DE | Wrapper (ELM) |
| Zorarpaci et.al. [16] | Single Objective | DE + ACO | Weka J48 classifier |
| Srikrishna et.al[17] | Single Objective | Quantum DE | Wrapper (LR) |
| Lopaez et.al. [18] | Single Objective | DE-FS(pm) | Wrapper (SVM) |
| Al-ani [67] | Single Objective | DE +Wheel based strategy | Filter |
| Zhao et.al. [19] | Single Objective | Modified DE | Wrapper (SVM) |
| Hancer [20] | Multi Objective | DE | Filter(Fuzzy+Kernel) |
| Li et.al. [21] | Single Objective | DE | Wrapper (SVM) |
| Wang et.al. [22] | Single Objective | DE | Wrapper (KNN) |
| Krishna and Ravi [68] | Single Objective | Adaptive DE | Wrapper (LR) |

Along with DE, several other evolutionary algorithms [23–27] considered the FSS problem like a combinatorial optimization problem. Khammassi and Krichen [28] proposed two schemes, namely, (i) The NSGA-BLR approach to handle binary-class datasets and (ii) The NSGA-MLR to handle multi-class network intrusion datasets. Chaudari and Sahu [29] proposed a binary version of the popular crow search algorithm (CSA) with time-varying flight in wrapper version BCSA-VF. Binary Dragon Fly algorithm (BDA) is proposed by Too and Mirjalili [30] by taking the Covid-19 dataset as a case study. Several variants of DE are employed, for example, in the estimation of tool-wear during the milling process [31], optimal resource scheduling [32], energy-efficient model [33], and anomaly detection [34].

Table 2 Parallel and distributed versions of DE and its variants

| Authors | Algorithm | Environment | Problem solved |
|---|---|---|---|
| | | | |

| Zhou[39] | DE | Spark | Pros and cons of various approaches is discussed |
| --- | --- | --- | --- |
| Teijeiroet.al. [40] | DE | Spark + AWS | Tested on benchmark functions |
| Chou et.al.[31] | DE | Spark | Clustering |
| Al-Sawwa and Ludwig[42] | DE | Spark | Designed a DE based classifier |
| Chen et.al.[43] | Modified DE | SPMD | Cluster Optimization |
| Adhianto et.al.[44] | DE | OpenMP | Optical Network problem |
| Liu et.al. [69] | DE | Distributed Cloud | Power electronic circuit optimization |
| Deng et.al. [45] | DE | Spark | Tested on benchmark functions and reported speedup |
| Wong et.al. [47] | Self-Adaptive DE | CUDA | Tested on benchmark functions and reported speedup |
| He et.al. [46] | Five variants of DE | Spark + Cloud | Developed a ring topology model and evaluated on benchmark functions to report speedup |
| Cao et.al. [48] | DPCCMOEA | MPI | Developed co-evolutionary based DE to solve large scale optimization |
| Ge et.al. [49] | DDE-AMS | MPI | Developed adaptive population model to solve large scale optimization |
| Falco et.al.[50] | DE | MPI | Resource allocation |
| Veronse and Krohling [51] | DE | CUDA | To solve large scale optimization in GPU environment |
| Glotik et.al. [52] | PSADE | MATLAB | Hydro Scheduling algorithm |
| Thomert et.al. [55] | NSDE-II | OpenMP | Cloud work placements |
| Daoudi et.al.[53] | DE | Hadoop | Clustering |
| Kromer et.al. [54] | DE | Unified Parallel C | To solve large scale optimization problems. |

Now we shift our attention to the works more relevant to the current study. Several parallel and distributed versions of the evolutionary algorithms [35–38] are proposed to handle high-dimensional datasets and big data. Various parallel and distributed implementations of the DE are presented in Table *2*. Zhou [39] discussed various strategies for implementing parallel DE MapReduce versions and their pros and cons in the Hadoop distributed framework. Teijeiro et al. [40] designed parallel DE under Spark environment, and the experiments were conducted in AWS cloud environment to solve benchmark optimization problems. Recently, Cho et al. [41] designed a parallel version of DE to solve large-scale clustering problems. Another parallel version of DE Classifier (SCDE) was proposed by Al-Sawwa and Ludwig [42] to handle the imbalanced data. SCDE finds the optimal centroid and assigns the class to the data point based on the Euclidian distance. Chen et al. [43] proposed a parallel version of the modified DE using single-program multiple-data (SPMD), with the improved genetic operators. They employed both fine-grained and coarse-grained approaches for cluster optimization. Adhianto et al. [44] proposed a fine-grained parallel version of DE using OpenMP to solve the optical networking problem, where the shared memory multi-processing is supported. Deng et al. [45] proposed a parallel DE for solving the benchmark functions and reported the speedup. He et al. [46] proposed the parallel framework for five variants of DE under the Spark cloud computing platform. They had analyzed the speedup by solving the benchmark functions. Wong et al. [47] developed the Computed Unified Device Architecture (CUDA) based framework for self-adaptive DE for solving the benchmark functions. Cao et al. [48] proposed a message passing interface (MPI) based co-evolutionary version of DE, where the population is divided and co-evolved together to solve large-scale optimization problems. Ge et al. [49] proposed an adaptive merge and split strategy for DE, namely, DE-AMS using MPI, to improve resource utilization, which is a vital aspect to minimize while handling large-scale optimization problems. Falco et al. [50] designed MPI-based DE under a CUDA grid environment and tested it on different resource allocation strategies. Veronse and Krohling [51] developed the first implementation of the CUDA version of DE. The proposed algorithm was tested on well known benchmark functions, and the computing time had been compared with the standalone implementation. Glotik et.al.[52] parallelized the DE using MATLAB to solve the hydro-scheduling problem. Daoudi et al. [53] developed the MapReduce version of DE under the Hadoop environment to solve the clustering problem. Kromer et al. [54] developed a parallel version of DE using Unified-C to handle large-scale clustering problems. Thomert et al. [55] developed a parallel version of DE using OpenMP to achieve the optimized workflow placement into the realm of practical utility.

The drawbacks in the extant approaches are as follows: (i) The existing EC based wrapper techniques, listed in Table 1, are limited to apply on small datasets and are sequential. Even though these can be applied on large datasets, the performance will be poor comapred to their parallel counterparts. (ii) Moreover, the current parallel and distributing EC techniques ( refer to Table 2) are not yet applied to FSS. These drawbacks motivated us to design scalable, parallel EC based wrapper techinques.

## 3. Overview of the methods employed

Evolutionary algorithms (EAs) are effective in getting global or near-global optimal solutions. The designing of EAs heuristics is inspired by natural selection and social behaviour. EA is an approach where the solutions are evolved throughout the process using Darwinian evolution principles. They start by initializing the population of solutions randomly. This population is evolved in order to determine better solutions. For the evolution to occur, EA's utilize specialized heuristics to generate new solutions and compute the corresponding fitness score given by the fitness function or (objective function). EA's are perfect for both continuous and binary search space.

### 3.1 Binary Differential Evolution

Differential Evolution (DE), a stochastic population-based global optimization algorithm, includes the heuristics, namely mutation, crossover, and selection. The mutation operation produces the mutant vector, which along with the candidate solution vector, undergoes crossover operation to generate the trail vector. At last, the selection operation is applied over the candidate solution vectors and trail vector to produce offspring. As mentioned earlier, this is continued till the completion of maximum iterations or other convergence criteria, if any, is met.

### 3.2 Binary Threshold Accepting

Dueck and Scheuer [56] proposed Threshold Accepting (TA) algorithm. Later, Ravi and Zimmermann [57] optimized a fuzzy rule-based classification model using modified TA. They developed the solution in three phases: feature selection was used as a preprocessing step, a modified fuzzy rule-based classification system was invoked over the selected feature subset, and finally, modified TA (MTA) was invoked to minimize the rule base size while guaranteeing high accuracy. Ravi et al. [58] proposed a modified TA (MTA) minimize the number of rules in a fuzzy rule-based classification system. Then,

Ravi and Zimmerman [59] proposed a continuous version of TA as an alternative to the backpropagation algorithm to overcome its limitations while training a neural network model. The trained neural network was utilized for feature selection, and the selected features were fed to the fuzzy classifier optimized by the MTA proposed in [58]. Later, Ravi and Pramodh [60] proposed principal component neural network architecture trained by TA for bankruptcy prediction of banks. Chauhan and Ravi [61] proposed a hybrid model DE with TA to solve unconstrained benchmark problems.

Threshold Accepting is a deterministic variant of simulated annealing. BTA, a binary version of the TA (named MTA in [58]), is presented in Algorithm 1. BTA performs neighbourhood search by flip-flopping the bits in the current solution vector one at a time, starting them in the left-most position. Each flip flop yields one neighbourhood solution. If the first neighbourhood solution is not accepted, then the bit is reversed to its original value. Likewise, 2nd bit is flip-flopped so on and so forth until a neighbourhood solution is accepted. However, it is not necessary to exhaustively search all the neighbourhood solutions. Searching a few of them would be sufficient, and that number is a user-defined parameter. The rest of the heuristics of the BTA are typical to that of the TA.

**Algorithm 1: Threshold Accepting (TA) algorithm**

```
1: Choose an initial configuration
2: Choose an initial THRESHOLD T > 0 and small number > 0
3: Opt: choose a new solution which is a small stochastic
   perturbation of the old solution.
4: compute AE:=
      fitness (new solution) – fitness (old solution)
5: IF AE < T THEN
6:    old solution := new solution
7:    IF in a long time, there is no increase in quality or
8:    too many iterations THEN
9:        Lower threshold by the equation T= T*(1- T)
10:   IF some time no change in quality anymore THEN stop
11:   GOTO opt
```

## 4. Proposed Scalable, Distributed, Parallel Wrapper

To perform FSS, we chose DE and TA, and built hybrids around them. DE explores and exploits the search space globally and is stochastic in nature. However, DE sometimes gets bogged down in local minima before convergence [32], thereby slowing down the convergence rate. If the search space becomes large, this phenomenon gets accentuated. Hence, it needs support from a local search-based optimization algorithm. Here, we chose to employ binary threshold accepting (BTA) for that purpose. The deterministic way of accepting the candidate solutions in BTA helps in exploration, exploitation and fast convergence.

Even though the EAs are intrinsically parallel, explicit parallel versions of EAs have to be designed so that they meet the following requirements: optimal utilization of the distributed resources, scalability, and low communication overhead. In general, the parallelism from the population perspective of EAs is achieved by two main models:
1. Master-slave (MS) strategy [40], which is also known as the global model. It has only a single global population. Here, the master takes the responsibility of applying metaheuristics (EAs subsumed) while the slave manages the fitness function evaluation.

2. The island strategy [40] is where the population is divided into islands, upon which the heuristics (operators) are applied independently.

The types mentioned above differ in the underlying topology and the migration rules to determine the communication between the nodes. A hybrid strategy can also be designed by combing the two strategies. In the present paper, we designed, MapReduce multithreaded framework, which mimics the combination of above-mentioned strategies.

The distributed framework like Apache Spark is not affected by the underlying topology because, in these frameworks, independent of the underlying topology, migrant solutions are broadcast to all the partitions.

The comparative analysis is carried out across PB-DE, PB-DETA, and PB-TADE to establish the importance of hybrid global and local search optimization heuristics. We designed parallel BTA, too, independent of the DE. If BTA alone is employed for FSS, then it consumed enormous computational time without yielding useful results. Hence, the comparative study excludes BTA. All the approaches follow the same solution encoding scheme and the population RDD encoding scheme as mentioned in section 4.1.

## 4.1 Wrapper Based FSS and solution encoding

Kohavi and John [71] are the pioneers in proposing the wrappers for FSS by posing the FSS as a combinatorial optimization problem. Here, wrappers take the help of a classifier or a regression model since it may involve the fitness score evaluation of a given feature subset.

### 4.1.1 Solution Encoding scheme

Wrappers using metaheuristics (EAs subsumed) require the fundamental step of solution encoding. EAs randomly initiate the population consisting of a set of solution vectors. Each solution vector in the population represents a feature subset. Such a solution vector comprises bits, where 1 indicates the presence, and 0 indicates the absence of a feature. The length/dimension of this array is equal to the number of features in the dataset.

### 4.1.2 Population RDD Encoding Scheme

Let the population, denoted by P, be initialized randomly using biased sampling from Uniform distribution between 0 and 1 as presented in Algorithm 2. Our objective is to select the less cardinal feature subsets yielding the highest AUC value. Hence, the parameter in the biased sampling is taken as 0.99. If the pseudo-random number is greater than 0.99, then the bit is assigned the value 1, indicating the feature's presence, and 0, otherwise, meaning the absence of the feature. Thus, according to Algorithm 2, the initialized population is taken as population RDD and the dataset as different RDD. The population RDD is presented in Table 3. The key is solution-id, and the first index of the value is the solution vector of binary type. It conveys which feature subset is selected. The second index stores the names or ids of the features, which helps us form the column-reduced dataset and the construction of pipeline RDD for the respective solutions. The last index stores the AUC corresponding to the solution vector.

**Algorithm 2: Biased sampling driven population initialization**

```
Input: n: population size, nfeat: number of features
Output: P: population RDD

1: While (i<n){
2:    While (j<nfeat){
3:        If rand() > 0.99:
4:            Feature is selected
5:            }
6:        }
```

Fig 1: Schema of the Population RDD

| Key | Value : Solution Vector |
|---|---|
| $Key_{\{1\}}$ | < Binary Vector$_{\{1\}}$, selected Features$_{\{1\}}$, AUC Scores$_{\{1\}}$ > |
| $Key_{\{2\}}$ | < Binary Vector$_{\{2\}}$, selected Features$_{\{2\}}$, AUC Scores$_{\{2\}}$ > |
| … | … |
| $Key_{\{n\}}$ | < Binary Vector$_{\{n\}}$, selected Features$_{\{n\}}$, AUC Scores$_{\{n\}}$ > |

## 4.2 Parallel Binary Differential Evolutionary (PB-DE)

This section discusses the proposed parallel binary differential evolution (PB-DE) for FSS under Spark. The PB-DE based wrapper algorithm is presented in Algorithm 3, and the flowchart of the execution flow is depicted in Appendix 1.

### 4.2.1 Population Initialization

The PB-DE algorithm starts by initializing the population according to Algorithm 2. Thus initialized population is stored in a different RDD and follows the structure as depicted in Fig.1. All the required parameters also get initialized. Then, the following information is broadcasted to all the nodes: mutation factor (*MF*), crossover rate (*CR*), number of features (*nfeat*), and population size (*n*). Once a variable is broadcasted, it is cached by the executor and utilized for other RDD operations, viz., transformations and actions. They are meant for read-only variables. By broadcasting, these variables are shared across the cluster, thereby reducing the communication overhead.

### 4.2.2 Fitness Score Evaluation on the Initial Population

Once the population is initialized, its fitness score is computed. As said earlier, the Population RDD is divided into different partitions, and each partition represents an island. These islands asynchronously undergo BDE heuristics as a separate thread in Spark. Using this *Binary vector information*, the corresponding selected features are obtained and updated in the *selected feature* column of the population RDD. Every time there is a change in the binary vector, the id information also gets updated. EA spends most of the time computing the fitness value. Hence, adopting an asynchronous way of fitness score (namely, AUC) calculation is the major task. This is achieved by initiating the thread pool mapper, where the number of threads is equal to the number of worker nodes. One cannot create the number of threads arbitrarily. Hence, after rigorous trials, the ideal size of the thread pool is found to the number of worker nodes. The size of the thread pool also affects the speedup. A low size thread pool leads to poor performance, whereas a higher thread leads to huge communication overhead.

Each thread in the thread pool is responsible for computing the AUC of the population island. Once this map is called, the reduced dataset is obtained based on the feature subset information using the *selected feature* index of each solution. Then, the ML pipeline is constructed where the reduced dataset and LR model are bound together, thereby giving pipelined RDD. Such a generated pipelined RDD is a subclass of the RDD, an immutable and partitioned collection of elements where all the operations are executed in parallel. LR model is trained by creating the vector assembler that has to be created for the corresponding selected features. Thus, vector assembler is created over the obtained reduced data frame. This kind of vector assembler is created for every solution in each iteration. As we have adopted the pipelined RDD, the above operations are executed in parallel and distributed across the nodes, thereby achieving fine-grained parallelism. Later, the AUC is evaluated with the training dataset in the same thread. All this process is repeated for each thread in the fitness value computation step. By adopting this strategy, the computation of AUC is performed in an asynchronous fashion. The same strategy is employed in all the proposed approaches to parallelize the fitness function evaluation. Thus, the AUC's are updated in the population RDD accordingly. This serves as the parent population. All the subsequent iterations follow the same thread pool mapper strategy to evaluate AUC.

### 4.2.3 Training phase

Using the binary vector field information the population undergoes DE heuristics, viz., Crossover and Mutation and forms the offspring population following the structure depicted in Fig.1. Using the Binary Vector information of thus formed offspring, their corresponding selected feature is also updated. Then the thread pool mapper is called, where the LR model evaluates the AUC on the reduced datasets. With the obtained AUC of each solution, their corresponding AUC column is also updated. Selection operator is applied on both the parent and offspring populations and the worst parent with less AUC is replaced with the better offspring with higher AUC. These better solutions are formed as one population, which serves as the parent population for the next iteration. This process of computing AUC and selecting the better solutions, thereby achieving the evolution of the population, is repeated for *maxIter* iterations.

### 4.2.4 Test phase

Thus evolved population obtained after *maxIter* is evaluated on the test dataset in the test phase. Here also the AUC is obtained by using a thread pool mapper only. Later, the evolved population is nothing but the set of the required feature subsets along with their corresponding test AUC.

**Algorithm 3: Proposed PB-DE based wrapper**

```
Input: P: population RDD, X: Input Dataset RDD, maxIter: maximum
iterations, MF: Muation Factor, CR: Cross over rate, n: population
size, nfeat: number of features

Output: P: Evolved population after maxIter

1: i← 0;
2: Initialize the population by biased sampling driven approach and
create population RDD; //Population initialization
3: Divide the population into partitions, each partition serves as an
   Island.
4: Create a thread pool of size number of nodes in the cluster;
     // Fitness score evaluation on Initial population
5: Evaluate the fitness function, each thread is responsible to train
LR model and evaluate the AUC for a solution in the population;
6: Broadcast the variables MR, CR, n, nfeat;

7: While(i< maxIter){ // Training phase
8:    Each sub population undergoes DE heuristics via BDE Mapper;
9:    Create a thread pool of size number of nodes in the cluster;
10:   Evaluate the fitness function, each thread is responsible to
      train LR and evaluate the AUC for every solution in the
      population.
11:   Replace worst parent solutions with best children solutions;
12:   Update the population RDD;
13:   Increment i ← i+1;
14: }
15:   Evaluate on Test dataset // Test  phase
16:   return Population, test AUC.
```

## 4.3 Parallel BDETA (PB-DETA)

It is important to note that the sequential DETA, which is parallelized and implemented in a distributed environment, is quite distinct from the DETA presented in Chauhan and Ravi [61]. The latter was a loosely–coupled hybrid system, wherein after DE converged, TA was invoked, whereas the current version of DETA is a tightly coupled system, in that DE and TA are invoked in tandem in every iteration and the hybrid algorithm is run for *maxIter1* iterations. Further, in the former, DE and TA were independently for many iterations, whereas we employed relaxed convergence criterion viz. running the two algorithms in tandem for only *maxIter1* iterations. This is strategy is designed to reduce computational time, primarily because we deal with big data sets in this paper.

The proposed parallel approach PB-DETA is presented in Algorithm 4, the schematic representation is depicted in Fig.2, and the flow chart of the execution flow is depicted in Appendix 2.

### 4.3.1 Population Initialization

PB-DETA algorithm also follows the same population initialization strategy of PB-DE and follows the structure as depicted in Fig.1. All the initialized parameters such as mutation factor (*MF*), crossover rate (*CR*), number of features (*nfeat*) and population size (*n*) are broadcasted.

### 4.3.2 Fitness Score Evaluation on the Initial Population

As explained earlier, EA algorithms spend most of the time evaluating the fitness scores. Hence, the same strategy which is used to parallelize the PB-DE evaluation phase is also used here. The fitness evaluation is done in parallel and asynchronously by creating a thread pool of the size equal to the number of nodes in the cluster.

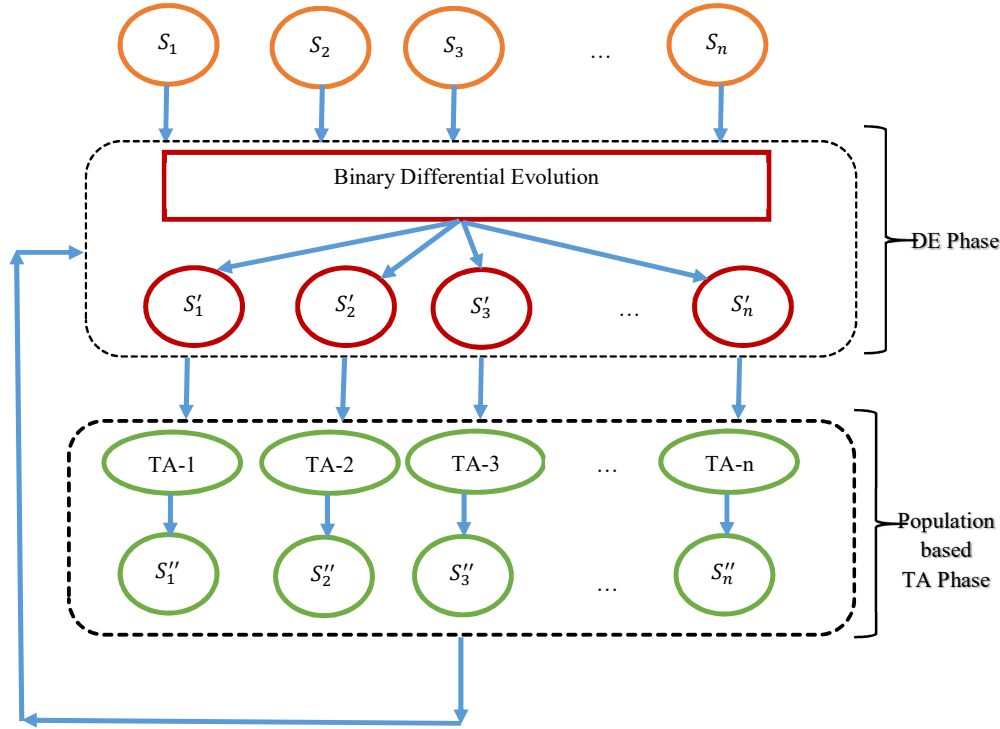

Fig. 2 Schematic representation of the DETA based wrapper

### 4.3.3 Training phase

Using the binary vector field information, the parent population undergoes BDE heuristics, viz, crossover, and population, forming the offspring population. Thread pool mapper is called on the offspring population to evaluate the AUC thereafter and the corresponding fields are updated. Then, the better offspring solutions replace the parent solutions. Thus evolved population undergoes BTA heuristics as given in Algorithm 2. Then, the thread pool mapper is called on, thus forming the offspring population to evaluate AUC, and also the corresponding fields are also updated. Here is the offspring solution, which is not much worse as per the threshold limit value, replaces the solution in the parent population. BTA heuristics are invoked for *maxIter2* times. Thus emerged population after *maxIter2* times serves as parent population.

After this, the above whole process of BDE and BTA in tandem is repeated until *maxIter1* iterations are completed.

*Algorithm 4: Proposed parallel* **PB-DETA** *based wrapper*

```
Input: P: population RDD, X: Input Dataset RDD, maxIter1: maximum DE
iterations, maxIter2: maximum TA iterations, MF: Mutation Factor, CR:
Cross over rate, n: population size, nfeat: number of features, T:
Threshold rate, eps: steps by which epsilon is decreasing after each
iteration

Output: P: Evolved population after maxIter1

1:  i ← 0;
2:  j ← 0;
3:  T ← 0.05;
4:  eps ← 0.95;
5:  Initialise the population by using biased sampling driven
    approach and create population RDD; //Population initialization
6:  Divide the population into partitions, each partition serves
    as an island;
     // Fitness score evaluation on Initial population
7:  Create a thread pool of size number of nodes in the cluster;
8:  Evaluate the fitness function, each thread is responsible to
    evaluate the AUC for a solution in the population;
9:  Broadcast the variables MR, CR, n, nfeat;

10: While(i< maxIter1){ // Training phase
11:    Each subpopulation undergoes DE heuristics via BDE Mapper;
12:    Create a thread pool of size number of nodes in the cluster;
13:    Evaluate the fitness function, AUC for a solution in the population;
14:    Replace worst parent solutions with best children solutions;
15:    Update the population RDD;
16:    Increment i ← i+1;
17:    j ← 0;
18:    While(j<maxIter2){
19:         Each subpopulation undergoes TA heuristics via BTA Mapper;
20:         Create a thread pool of size number of nodes in the cluster;
21:         Evaluate the fitness function, each thread is responsible
                to evaluate the AUC for a solution in the population;
22:         Replace  parent  solutions  with  not  much  worse  children
    solutions;
23:         Update the population RDD;
24:         Increment j ← j+1;
25:         eps ← eps*(1-W);
26:         }
27:    }
28:    Evaluate on Test dataset  // Test phase
29:    return Population, test AUC.
```

### 4.3.4 Test phase

Then, the test phase begins, where the obtained evolved population after maxIter1 number of iterations is evaluated on the test dataset. Here also the AUC is computed by using a thread pool mapper. Later, the so evolved population is nothing but the set of required feature subsets with their corresponding test AUC.

## 4.4 Parallel TABDE (PB-TADE)

The proposed parallel approach PB-TADE is presented in Algorithm 5, the schematic representation is depicted in Fig. 3, and the flow chart is depicted in Appendix 3.

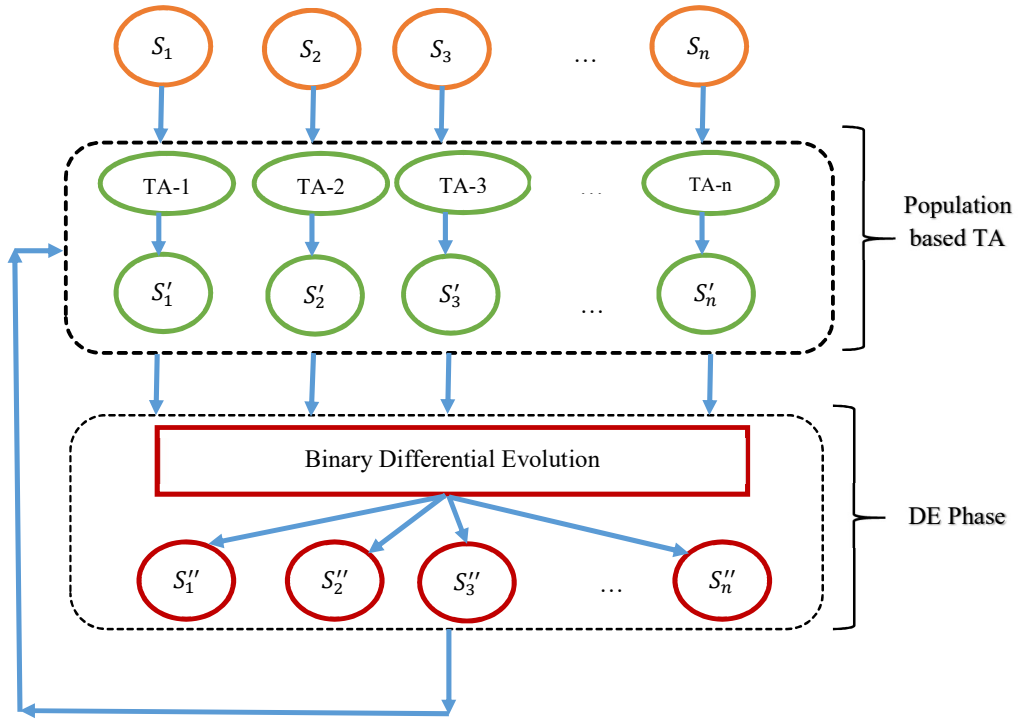

Fig.3: Schematic representation of the PB-TADE based wrapper

### 4.4.1 Population Initialization

PB-TADE algorithm also follows the same population initialization strategy of PB-DE and follows the structure as depicted in Fig.1. All the initialized parameters such as mutation factor (*MF*), crossover rate (*CR*), number of features (*nfeat*) and population size (*n*) are broadcasted.

### 4.4.2 Fitness Score Evaluation on the Initial Population

As explained earlier, EA algorithms spend most of the time evaluating the fitness scores. Hence, the same strategy which is used to parallelize the PB-DE evaluation phase is also used here. The fitness function is evaluated in parallel and asynchronously by creating a thread pool of the size equal to the number of nodes in the cluster.

**Algorithm 5: Proposed parallel PB-TADE based wrapper**

*Input* P: Population RDD, X: Input Dataset RDD, maxIter1: maximum iterations BDE invokes, maxIter2: maximum iterations BTA invokes, MF: Mutation Factor, CR: Crossover Rate, n: population size, nfeat: number of features, T: Threshold rate, eps: steps by which epsilon is decreasing after each iteration

*Output* P: Evolved Population after maxIter1

```
1:  i ← 0;
2:  j ← 0;
3:  T ← 0.05;
4:  eps ← 0.95;
5:  Initialise the population by using biased sampling driven approach
    and create population RDD; //Population initialization
6:  Divide the population into partitions, each partition serves as an
island;
// Fitness score evaluation on Initial population
7:  Create a thread pool of size number of nodes in the cluster;
8:  Evaluate the fitness function, each thread is responsible to evaluate
    the AUC for a solution in the population;
9:  Broadcast the variables MR, CR, n, nfeat;
10: While(i< maxIter1){ // Training phase
11:     j ← 0;
12:     While(j<maxIter2){
13:         Each subpopulation undergoes TA heuristics via BTA Mapper;
14:         Create a thread pool of size number of nodes in the cluster;
15:         Evaluate the fitness function, each thread is responsible
            to evaluate the AUC for a solution in the population;
16:         Replace parent solutions with not much worse children
            solutions;
17:         Update the population RDD;
18:         Increment j ← j+1;
19:         eps ← eps*(1 – W);
20:     }
21:     Each subpopulation undergoes DE heuristics via BDE Mapper;
22:     Create a thread pool of size number of nodes in the cluster;
23:     Evaluate the fitness function, each thread is responsible to evaluate
        the AUC for a solution in the population;
24:     Replace worst parent solutions with best children solutions;
25:     Update the population RDD;
26:     Increment i ← i+1;
27: }
28: Evaluate on Test dataset // Test phase
29: return Population, test AUC.
```

### 4.4.3 Training phase

This algorithm is the hybrid of BTA and BDE. Here, BTA is executed first for maxIter2 times, followed by BDE. It means that first, the local exploitation happens in each solution individually. Later, it is followed by global exploration and exploitation over the entire population.

Using the binary vector field information, the parent population undergoes BTA heuristics as presented in Algorithm 2 and forms the offspring population. Then, AUC is evaluated by calling the thread pool mapper on the offspring population. Thus evolved population after undergoing BTA for maxIter2 times forms a population which serves as the initial parent population for BDE. This population undergoes the BDE heuristics, viz, crossover and mutation. After that, the thread pool mapper is called and the AUC of the offspring population is evaluated. Then, the selection operation is applied thereby replacing the worse parent solutions with their better offspring solutions. This serves as the parent population for the next iteration.

After this, the whole process of BTA and BDE in tandem is repeated until maxIter1 iterations are completed.

### 4.4.4 Test phase

Then the test phase begins where the population obtained at the end of the training phase, is evaluated on the test dataset. Here also the AUC is evaluated by using a thread pool mapper only. Later, gives the so evolved population is nothing but the set of the required feature subsets with their corresponding test AUC.

### 4.5 Classification Algorithm

Logistic Regression (LR) is employed as the classifier for the proposed FSS. LR is chosen because it is fast to train and is nonparametric. It does not make any assumptions about the errors or variables, takes less time to converge, and has no hyperparameters to fine-tune.

### 4.6 Fitness Function

Area under the ROC Curve (AUC) is the fitness function for our proposed wrappers. It is proven to be robust measure than accuracy for unbalanced datasets. It is defined as the average of specificity and the sensitivity. The cut-off for the classification of probability in binary classes is taken as 0.5. Accordingly,

$$AUC = \frac{Sensitivity + Specificit}{2} \quad (7)$$

where, sensitivity is the ratio of the positive samples that are correctly predicted to be positive to all the predicted positive samples. This is also called True Positive Rate (TPR).

$$Sensitivity = \frac{TP}{TP+FN} \quad (8)$$

where, TP is true positive and FN is false negative and specificity is the ratio of the negative samples that are correctly predicted to be negative to all the predicted negative samples. This is also called True Negative Rate (TNR).

$$Specificity = \frac{TN}{TN+FP} \quad (9)$$

where, TN is true negative and FP is false positive. These are obtained from the confusion matrix.

## 5 Dataset Description and Experimental setup

The meta-information of the benchmark datasets is presented in Table 3. All other datasets except for the Microsoft Malware dataset, contain categorical features. Thus, categorical features are handled by using one-hot encoding mechanism. All the datasets pertain to binary classification problems. The Microsoft Malware dataset is accessed from the Kaggle repository [74], IEEE malware dataset from IEEE data port [75], whereas OVM_Omentum and uterus are genomic datasets from open source OpenML datasets [77], and the Epsilon dataset from LIBSVM binary dataset repository [76].

All the experiments are conducted in a Spark-HDFS cluster with Spark version 2.4.4 and Hadoop version 2.7, having one master node and four worker nodes with 32 GB RAM with Intel i5 8th generation.

Table 3: Description of the benchmark datasets

| Name of the Dataset | # Objects | # Features | # Classes | Size of the Dataset |
|---|---|---|---|---|
| Epsilon | 5,00,000 | 2000 | 2 | 10.8 GB |
| Microsoft Malware | 32,59,724 | 76 | 2 | 1.8 GB |
| IEEE Malware | 15,00,000 | 1000 | 2 | 3.2 GB |
| OVM_Omentum | 1584 | 10,935 | 2 | 108.3 MB |
| OVM_Uterus | 1584 | 10,935 | 2 | 108.3 MB |

## 6 Results & Discussions

All the datasets are divided into training and test sets in the ratio 80%:20%. Stratified random sampling is performed to maintain the similar proportion of the classes in the training and test datasets. It is well-known that the performance of the EC techniques is susceptible to change in hyperparameters. Hence, after rigorous fine-tuning with several combinations, the hyperparameters are frozen and listed in Table 4. For each algorithm and for datasets, the population size is fixed at 10, and the maximum number of generations is taken as 20. In the case of PB-DE, the DE is executed for 20 generations. However, in the case of the two hybrids (PB-DETA and PB-TADE) DE and TA are individually executed for 10 generations each, thereby making it 20 generations in all. All the experiments are repeated for 20 runs to nullify the impact of the random seed, which is customary for all evolutionary computing techniques. The top solutions that achieved the highest AUC in each run are considered to report the average highest AUC and the corresponding average cardinality over 20 runs (see Table 5).

Table 4: Hyperparameters for all the approaches

| Dataset | PB-DE | | PB-DETA | | PB-TADE | |
|---|---|---|---|---|---|---|
| | MF | CR | MF | CR | MF | CR |
| Epsilon | 0.8 | 0.8 | 0.8 | 0.8 | 0.8 | 0.8 |
| Microsoft Malware | 0.8 | 0.9 | 0.8 | 0.9 | 0.8 | 0.9 |
| IEEE Malware | 0.8 | 0.9 | 0.8 | 0.9 | 0.8 | 0.9 |
| OVM_Omentum | 0.75 | 0.9 | 0.75 | 0.9 | 0.75 | 0.9 |
| OVM_Uterus | 0.85 | 0.9 | 0.85 | 0.9 | 0.85 | 0.9 |

## 6.1 Three-way comparative analysis

The 3-way comparison is conducted on the performance of the three models to establish the importance of proposed hybrid approaches over the PB-DE.

The results in Table 5 show that PB-TADE can achieve the best AUC because PB-DE got stuck in the local minima. Both PB-DETA and PB-TADE avoided this as they have employed with TA either before or after DE. The advantage of finding the local search exploitation helps not to get entrapped in the local minima but also find the better maxima. The feature subsets selected by PB-DE achieved less accuracy when compared to both PB-DETA and PB-TADE. Also, the average cardinality obtained by PB-DE is relatively high compared to that of both the PB-DETA and PB-TADE. Both these cases are not ideal in obtaining an optimal solution. Invocation of TA is also necessary while designing the hybrid model. Hence, we worked on both the possibilities as part of the ablation study. We designed parallel BTA, too, independent of the DE. If BTA alone is employed for FSS, then it consumed enormous computational time with inferior results. Hence, the comparative study excludes BTA.

In the ablation experiment involving PB-TADE and PB-DETA, the former achieved little lesser average cardinality of the feature subsets than the latter. In addition to that, the mean AUC is a little less but quite comparable in the case of IEEE Malware and Microsoft Malware. The former outperformed the latter because (i) BTA is essentially very good at local search by virtue of it being a point-based algorithm and a deterministic variant of simulated annealing (ii) even though we proposed a population-based BTA, in these hybrids, the hallmark of population-based evolutionary algorithm namely passing on the knowledge learned by the individual solutions to one another from generation to generation is conspicuously missing by design. Therefore, they are at the most population size number of BTA instances running independently. (iii) Thus, in PB-TADE, after BTA does the exploitation well, the baton is passed on to the BDE, which is demonstrably superior in both exploration and exploitation. This cycle continues in every iteration. (iv) However, in PB-DETA, the BDE does the job of exploration and exploitation well before the BTA is invoked, which only minimizes the fitness values obtained by DE. Further, we should note that both BTA and BDE are run for 10 iterations each, which means that they are run with relaxed convergence criteria without adversely impacting the fitness value or the AUC. This is a significant departure from the traditional implementations of both BDE and BTA for solving combinatorial optimization problems, where they are typically run for many iterations for convergence. This strategy is designed to reduce the computational time, primarily because we deal with big data sets in a distributed manner in this paper.

Table 5: Average Cardinality and mean AUC obtained

| Dataset | PB-DE | | PB-DETA | | PB-TADE | |
|---|---|---|---|---|---|---|
| | Average Cardinality | Mean AUC | Avg. Cardinality | Mean AUC | Avg. Cardinality | Mean AUC |
| Epsilon | 617.3 | 0.7932 | 486.1 | 0.8029 | 457.7 | **0.8098** |
| Microsoft Malware | 29.6 | 0.6872 | 21.7 | 0.7002 | 18.60 | **0.7054** |
| IEEE Malware | 643.45 | 0.7929 | 477.9 | 0.8035 | 463.9 | **0.8109** |
| OVA_Omentum | 47.28 | 0.8607 | 35.54 | 0.8722 | 26.15 | **0.8817** |
| OVA_Uterus | 37.3 | 0.8607 | 28.60 | 0.8712 | 27.12 | **0.8802** |

Further, no feature subset selection work is reported in analyzing Microsoft Malware and IEEE Malware datasets to the best of our knowledge. In the Epsilon dataset, Peralta et al. [78] designed MapReduce for evolutionary feature selection. They used CHC as the evolutionary strategy, logistic regression as the classifier and achieved a 0.6985 AUC with 721 features, whereas PB-TADE obtained an average AUC of 0.8098 with 457.7 average number of features. Moreover, Pes [79] conducted feature selection by using Symmetric Uncertainty (SU), while AUC the scores are computed using Random Forest (RF). They reported an AUC of 0.695 and 0.6 in the OVA_Uterus and OVA_Omentum datasets, respectively. However, they did not report the optimal number of features that obtained these scores. But, PB-TADE achieved an average AUC of 0.8802 and 0.8817. In comparison, the OVA_Uterus and OVA_Omentum datasets have an average number of features, 27.12 and 26.15, respectively. Thus, our proposed methods outperformed the state-of-the-art results in these datasets.

## 6.2 Repeatability

Repeatability is one of the critical criteria for how robust and stable the designed wrapper method is. The more an optimal feature or feature subset repeats itself, the more powerful the underlying evolutionary algorithm is said to be. In this subsection, repeatability analysis is conducted in two ways. Firstly, concerning the features repeated individually among the often-repeated feature subsets with the highest AUC and secondly, the repetition of a feature subset as a whole corresponding to the highest AUC.

### 6.2.1 Repeatability of the Individual Features

All the most repeated features part of an optimal feature subset with the highest AUC are identified and presented in Table 6. The features repeated for more than 50% of the total individuals obtained by 20 runs are considered and presented. Results accommodate the most repeated top five features selected by each approach. It turns out that the repeated features selected by PB-DETA and PB-TADE are almost identical whereas, the features chosen by the PB-DE are slightly different.

Table 6: Most repeated features selected by each approach

| Dataset | Approach | Most repeated features |
|---|---|---|
| Epsilon | PB-DE | 1,3,5,7,9 |
| | PB-DETA | 1,3,6,12,19 |
| | PB-TADE | 1,3,6,12,19 |
| Microsoft Malware | PB-DE | AVProductsInstalled,HasTpm,Isprotected,Census_OEMN_Name Identifier,SmartScreen |
| | PB-DETA | AVProductsInstalled,HasTpm,IsPassiveMode, OsSuite,SmartScreen |
| | PB-TADE | AVProductsInstalled,HasTpm,OsSuite, RipStateBuild,SmartScreen |
| IEEE Malware | PB-DE | GetProcAddress,GetThreadId,Sleep,FindClose, RaiseException |
| | PB-DETA | GetProcAddress,GetLastError,Sleep,ReadFile, RaiseException |
| | PB-TADE | GetProcAddress,GetLastError,Sleep,ReadFile, RaiseException |
| OVA_Omentum | PB-DE | 158765_at,201608_s_at, 206442_at,207096_s_at,210002_s_at |
| | PB-DETA | 1554436_s_at, 201669_s_at, 20644_s_at, 207442_s_at, , 208970_s_at |
| | PB-TADE | 1554436_s_at, 201669_s_at, 20644_s_at, 207442_s_at, , 208970_s_at |
| OVA_Uterus | PB-DE | 205866_s_at,209682_s_at,217294_s_at, 222421_s_at,220148_s_at, |
| | PB-DETA | 202125_s_at,205866_s_at,218132_s_at, 222421_s_at,222784_s_at, |
| | PB-TADE | 202125_s_at,205866_s_at,218132_s_at, 222421_s_at,222784_s_at, |

### 6.2.3 Repeatability of the Feature Subsets

All the feature subsets that yielded the highest AUC and repeated often are reported in Table 7. The #s1 represents the cardinality of the most-repeated feature subset, and the corresponding AUC. The cardinalities of the most repeated top most feature subset is presented in Table 7. In the case of the Epsilon dataset, PB-DE has selected 639 features resulting in 79.67%. PB-TADE outperformed the PB-DE in terms of AUC by selecting a less cardinal feature subset. In the Microsoft Malware dataset, PB-DE achieved a 69.74% AUC with 31 features, while both PB-DETA and PB-TADE can achieve a better AUC than PB-DE with a less cardinal feature subset. Even though in the case of the IEEE Malware dataset, the cardinality of the selected feature subsets are the same for the PB-DE, PB-DETA, and PB-TADE, the AUC are different because the selected features are not identical. The same is the case in OVA_Omentum and OVA_Uterus datasets. Hence, the results indicate that PB-TADE can achieve lower cardinality with better AUC than PB-DETA and PB-DE in all the datasets in terms of repeatability. Similarly, PB-DETA outperformed PB-DE.

Table 7: Cardinalities and the corresponding AUC of the Top-most repeated feature subsets

| Dataset | PB-DE | | PB-DETA | | PB-TADE | |
| --- | --- | --- | --- | --- | --- | --- |
| | #s1 | AUC | #s1 | AUC | #s1 | AUC |
| Epsilon | 639 | 0.7967 | 564 | 0.8014 | 488 | **0.8097** |
| Microsoft Malware | 31 | 0.6974 | 24 | 0.7001 | 17 | **0.7061** |
| IEEE Malware | 550 | 0.7956 | 486 | 0.8057 | 487 | **0.8058** |
| OVA_Omentum | 55 | 0.8701 | 37 | 0.8723 | 31 | **0.8779** |
| OVA_Uterus | 39 | 0.8598 | 42 | 0.8723 | 32 | **0.8850** |

* where #s1 is the cardinality of the top-most repeated feature subset

## 6.3 Least Cardinal Feature Subset with highest AUC

In this subsection, the least cardinal feature subset among the most repeated feature subset with the highest AUC is discussed. The results are presented in Table 8. It turns out that the PB-TADE outperformed PB-DETA and PB-DE in detecting the most repeatable least cardinal feature subset, while PB-DETA stands second in detecting the most repeatable least cardinal feature subset. Except for OVA_Omentum dataset, PB-TADE achieved the better AUC with lesser number of features than the PB-DETA and PB-DE. In the case of OVA_Uterus, PB-DETA achieved better AUC with lesser number of features than the PB-DE. In all the cases, PB-DE is outperformed by the proposed hybrid models. This fact further reinforces the role played by the BTA in this hybridization.

Table 8: Least Cardinal Feature subset selected by each approach with the most repetitions

| Dataset | PB-DE | | PB-DETA | | PB-TADE | |
| --- | --- | --- | --- | --- | --- | --- |
| | #s1 | AUC | #s1 | AUC | #s1 | AUC |
| Epsilon | 588 | 0.7967 | 471 | 0.8068 | 487 | **0.8097** |
| Microsoft Malware | 27 | 0.6915 | 22 | 0.7007 | 17 | **0.7061** |
| IEEE Malware | 550 | 0.7956 | 484 | 0.8057 | 487 | **0.8197** |
| OVA_Omentum | 41 | 0.8504 | 29 | **0.8723** | 33 | 0.8699 |
| OVA_Uterus | 37 | 0.8504 | 37 | 0.8723 | 32 | **0.8750** |

* where #s1 is the cardinality of the feature subset having least cardinal subset with highest AUC

## 6.4 Speedup

Speedup is defined as the gain obtained by the parallel version of the algorithm with respect to the sequential algorithm executed on a single processor as follows in Eq.(11)

$$\text{Speed Up (S.U)} = \frac{\text{Time taken by Sequential Version}}{\text{Time taken by Parallel Version}} \quad (11)$$

Table 9: Speedup Analysis of parallel versions over the sequential ones

| Dataset | BDE E.T (sec) | PB-DE E.T (sec) | S.U | DETA E.T (sec) | PB-DETA E.T (sec) | S.U | TADE E.T (sec) | PB-TADE E.T (sec) | S.U |
|---|---|---|---|---|---|---|---|---|---|
| Epsilon | 12780 | 4485 | **2.84** | 12680 | 4361 | **2.90** | 12688 | 4369 | **2.90** |
| Microsoft Malware | 16412 | 6741 | 2.43 | 15781 | 6447 | 2.44 | 15779 | 6432 | 2.45 |
| IEEE Malware | 20412 | 8151 | 2.50 | 19793 | 7936 | 2.49 | 19801 | 7938 | 2.49 |
| OVA_Omentum | 14892 | 5428 | 2.74 | 14651 | 5231 | 2.80 | 14689 | 5226 | 2.81 |
| OVA_Uterus | 14108 | 5378 | 2.62 | 13979 | 5207 | 2.68 | 13968 | 5222 | 2.67 |

* where E.T is the execution time given in seconds

This stands as one of the essential characteristics in evaluating the performance of the parallel version of the algorithm. The results are presented in Table 9. It is to be noted that speedup results are rounded off to two decimals. We observed that the proposed parallel algorithms achieved significant speedup. Speedup achieved ranges from 2.43 – 2.90 times by all proposed algorithms over their sequential counterparts in all datasets. As the number of nodes in the cluster are 4; the maximum possible speedup could be achieved is 4. The linear speedup is not achieved because of the synchronization junctions in the parallel model.

## 6.5 Statistical testing of the results

The two-tailed t-test at 5% level of significance at 38 (20+20-2) degrees of freedom is conducted pairwise on the three proposed algorithms to make statistically valid statements about their performance. The null hypothesis is $H_0$: *both the algorithms are statistically equal*, while the alternate hypothesis is, $H_1$: *both the algorithms are statistically not equal*.

Table 10: Pairwise Paired T-test analysis: A three way comparison

| Dataset | PB-DE vs. PB-DETA | | PB-DE vs. PB-TADE | | PB-DETA vs. PB-TADE | |
|---|---|---|---|---|---|---|
| | t-statistic | p-value | t-statistic | p-value | t-statistic | p-value |
| Epsilon | 4.0930 | 0.00021 | 7.72 | 2.66 x$10^{-09}$ | 3.56 | 0.0010 |
| Microsoft Malware | 3.318 | 0.002 | 4.62 | 4.25 x$10^{-05}$ | 3.10 | 0.0035 |
| IEEE Malware | 4.403 | 8.39x$10^{-05}$ | 8.045 | 9.91 x$10^{-10}$ | 3.63 | 0.0008 |
| OVA_Omentum | 4.81 | 2.36 x$10^{-05}$ | 4.168 | 0.00017 | 2.06 | 0.0453 |
| OVA_Uterus | 3.56 | 0.00099 | 3.69 | 0.00069 | 1.74 | 0.0891 |

The top accuracy scores achieved by each approach over the 20 runs are considered for the t-test evaluation. As the p-values for all datasets are less than 0.05, the null hypothesis is rejected, and alternate hypothesis is accepted. The t-statistic and p-value are reported in Table 10. We infer that PB-DETA and PB-TADE are significantly different from PB-DE, as the p-values are significantly small. Further, in all datasets except for the OVA_Uterus, the PB-TADE turned out to be statistically significantly better than PB-DETA.

## 7. Conclusion

This paper develops the parallel versions of the BDE, BDETA, and BTADE and employs them for the wrapper based feature subset selection, where logistic regression is chosen as the classifier. We demonstrated their effectiveness on five high-dimensional datasets taken from literature. The results indicate that the PB-TADE is statistically significant compared to PB-DETA and PB-DE at both exploration and exploration. All the proposed parallel approaches achieved significant speedup compared to their sequential counterparts. Although PB-TADE and PB-DETA consumed more time than the PB-DE, it is mainly due to more function evaluations. It is noteworthy that the resulting optimal feature subsets are much better than those of the PB-DE in terms of higher AUC and less cardinality. Further, our proposed methods outperformed the state-of-the-art results, wherever the results were reported. The above analysis is conducted in a single-objective function environment. In the future, the investigation will be carried out on the same problem but in bi-objective and multi-objective environments.

## Appendix

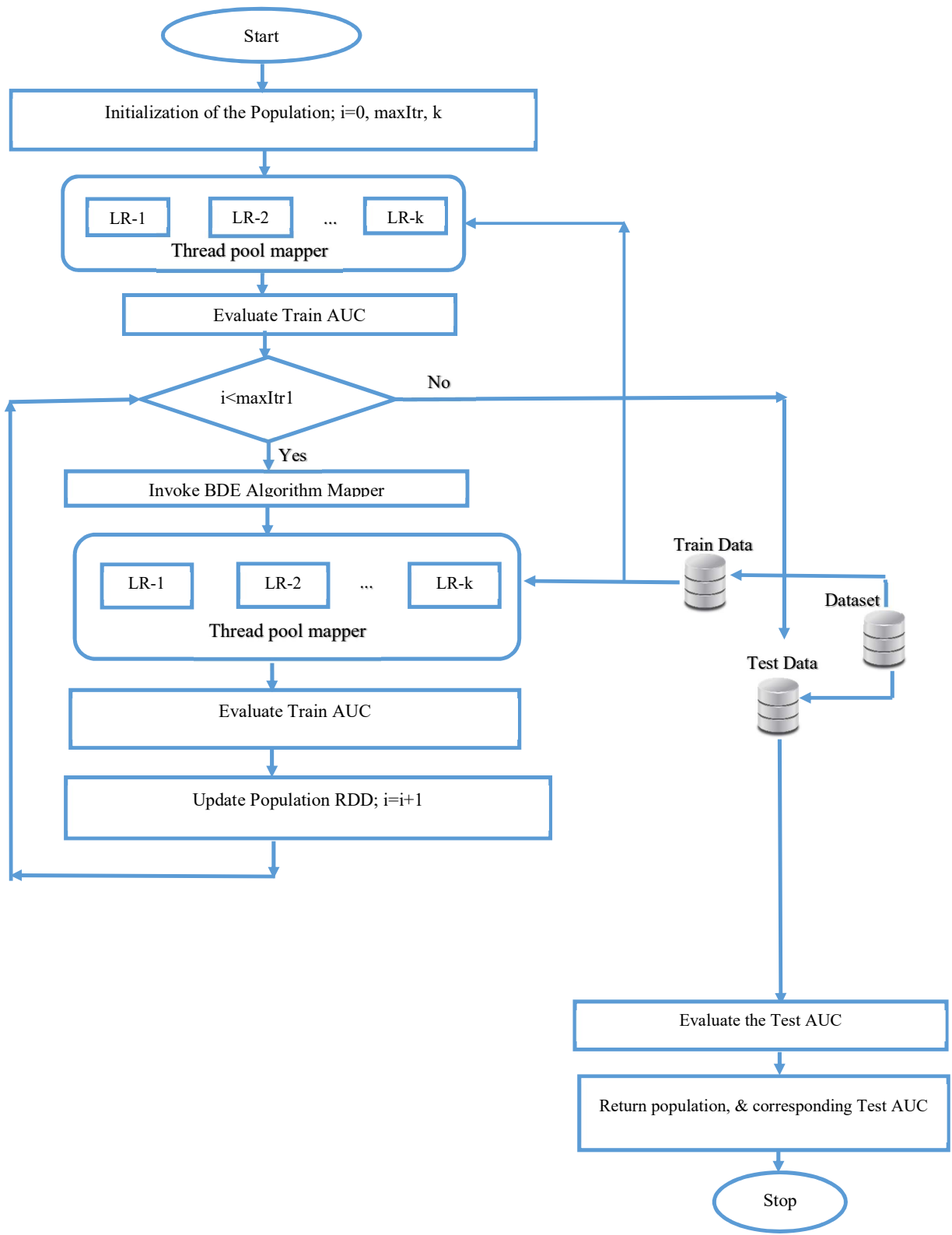

Fig.1: Flowchart of PB-DE based wrapper

V

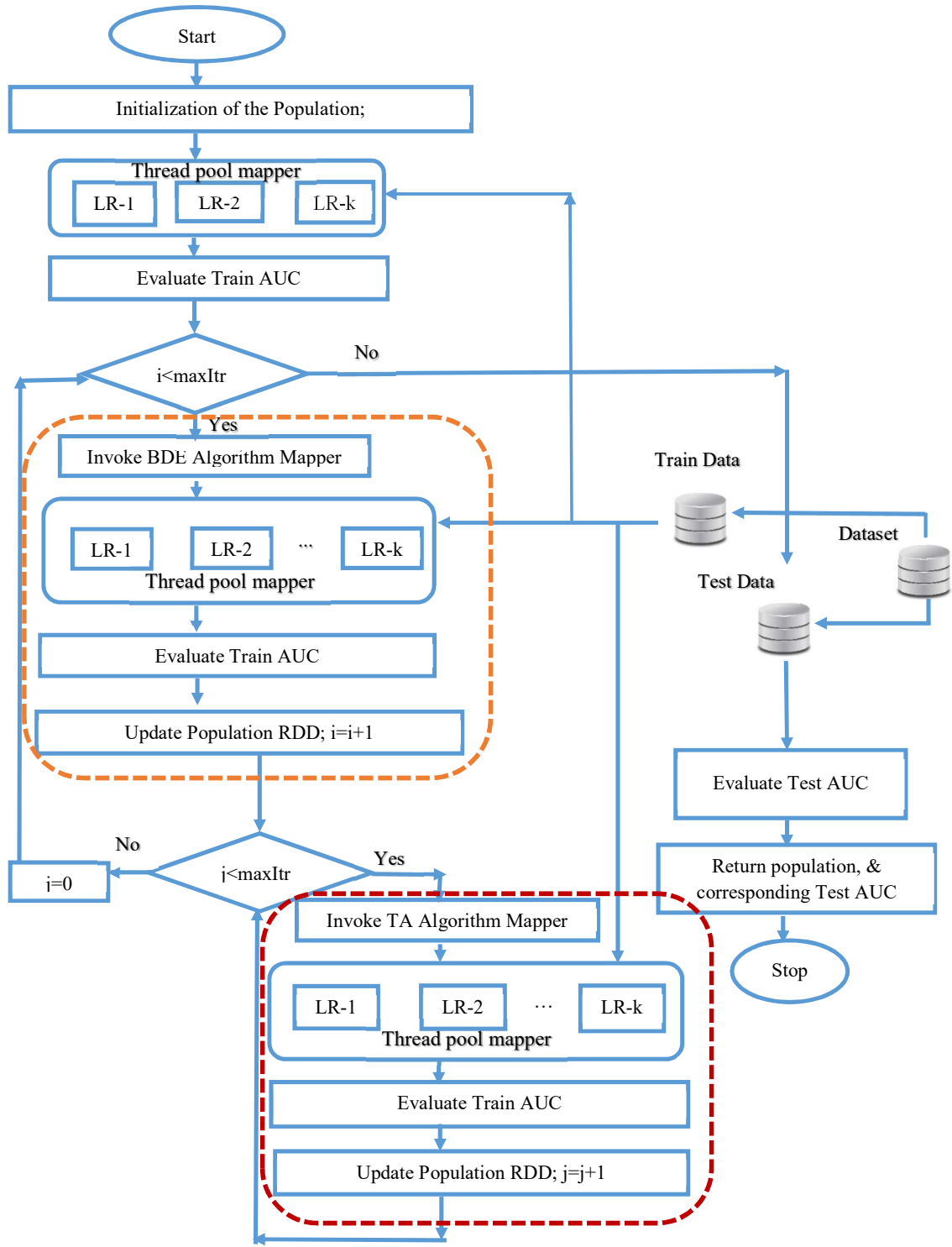

Fig.2: Flow chart of the PB-DETA based wrapper

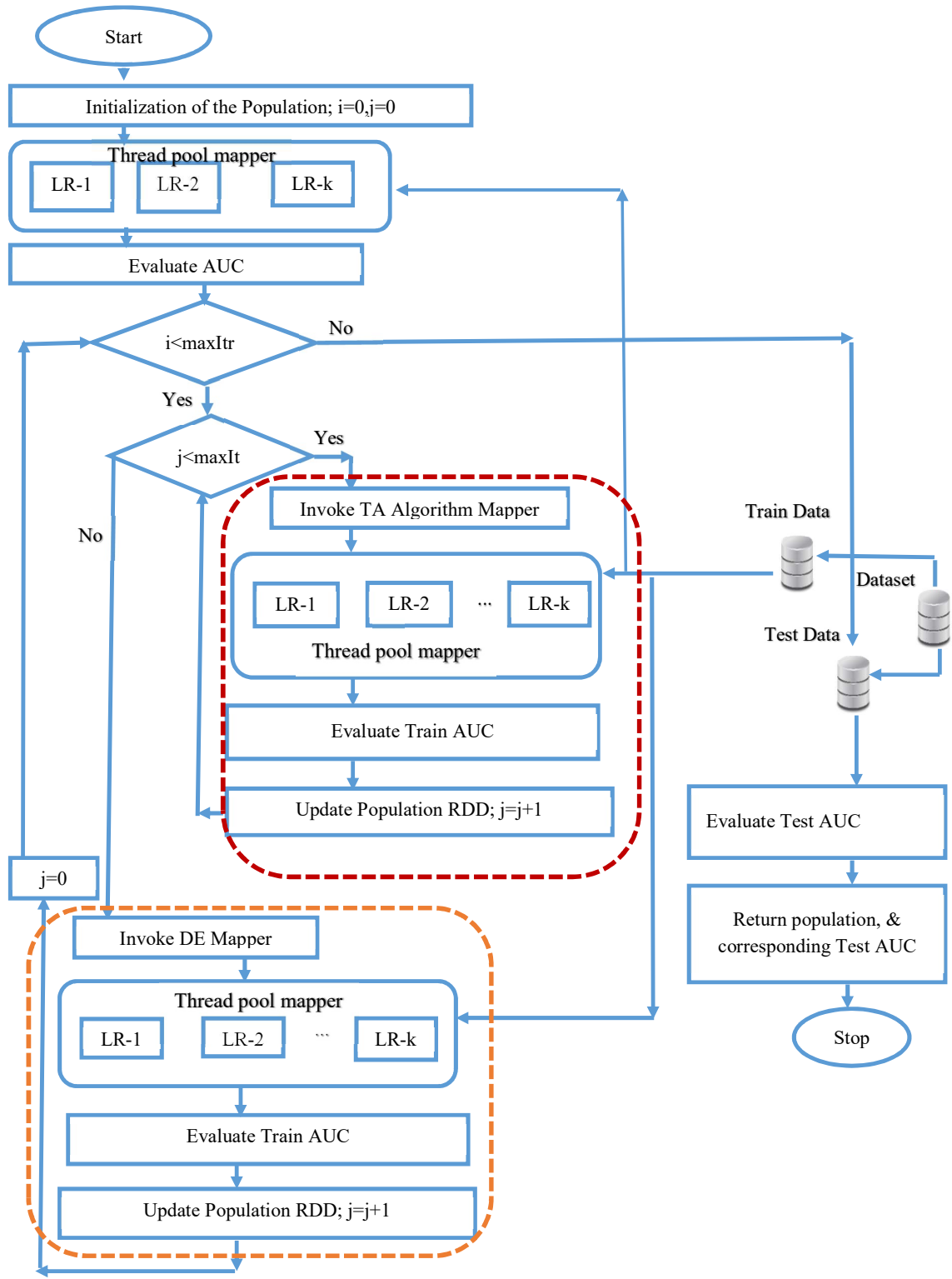

Fig. 3: Flowchart of the parallel PB-DETA based wrapper